# Multiscale lubrication simulation based on fourier feature networks with trainable frequency

TANG Yihu[1,2,3], HUANG Li[1,2,3], WU Limin[2,3], MENG Xianghui[1, *]

[1] *School of Mechanical Engineering, Shanghai Jiao Tong University, Shanghai, 200240, China*

[2] *Shanghai marine diesel research institute, Shanghai, 201108, China*

[3] *National key laboratory of marine engine science and technology, Shanghai, 201108, China*

**Abstract:** Rough surface lubrication simulation is crucial for designing and optimizing tribological performance. Despite the growing application of Physical Information Neural Networks (PINNs) in hydrodynamic lubrication analysis, their use has been primarily limited to smooth surfaces. This is due to traditional PINN methods suffer from spectral bias, favoring to learn low-frequency features and thus failing to analyze rough surfaces with high-frequency signals. To date, no PINN methods have been reported for rough surface lubrication. To overcome these limitations, this work introduces a novel multi-scale lubrication neural network architecture that utilizes a trainable Fourier feature network. By incorporating learnable feature embedding frequencies, this architecture automatically adapts to various frequency components, thereby enhancing the analysis of rough surface characteristics. This method has been tested across multiple surface morphologies, and the results have been compared with those obtained using the finite element method (FEM). The comparative analysis demonstrates that this approach achieves a high consistency with FEM results. Furthermore, this novel architecture surpasses traditional Fourier feature networks with fixed feature embedding frequencies in both accuracy and computational efficiency. Consequently, the multi-scale lubrication neural network model offers a more efficient tool for rough surface lubrication analysis.

**Keywords:** physics-informed neural network, Fourier feature embedding, multiscale lubrication, rough surface

## Nomenclature

| | |
|---|---|
| $\alpha$ | Inclination angle of the slider (deg) |
| $h$ | Local film thickness (μm) |
| $h0$ | Minimum film thickness (μm) |
| $L$ | Slider bearing length (mm) |
| $B$ | Slider bearing width(mm) |
| $H, H_0$ | Non dimensional film thickness (-) |
| $p$ | Pressure (Pa) |
| $P$ | non-dimensional pressure (-) |
| $u$ | The relative slide velocity of the surfaces (m/s) |
| $\eta$ | The viscosity of the lubricant (Pa.m/s) |
| $f$ | Feature frequency |
| $\varphi^i$ | Fourier feature mappings |
| $\sigma$ | Activation function |
| $\sigma_i$ | The fourier feature sampling deviation |
| $w$ | Network weight |
| $b$ | Network bias |
| $\Phi_i (x)$ | The ADF to each curve or line segment |

## 1 Introduction

Surface roughness plays a crucial role in optimizing contact interface performance, mitigating wear, and enhancing lubrication efficiency. In recent years, there has been a growing recognition of the significance of surface roughness, leading to a concerted effort in designing surface textures meticulously to mitigate the adverse effects of rough contact surfaces. The key to achieving this objective is to develop computational methodologies capable of accurately and efficiently modeling the intricate interplay between surface topography and lubrication performance.

Multi-scale numerical techniques offer a hybrid approach by integrating microscale and macroscale physical processes to enable cross-scale modeling of lubrication phenomena on rough surfaces[1]. Pei [2, 3]introduced an innovative finite element method that enhances simulations of surface texture effects in hydrodynamic lubrication. Niemec[4, 5] proposed a multiscale approach that integrates a microscale deterministic mixed lubrication models with macroscale models to accurately predict load-bearing changes due to surface textures. Brunetière[6, 7] developed a multiscale finite element method tailored for analyzing hydrodynamic lubrication problems on extensive rough contact surfaces.

Given the computational complexity inherent in multiscale issues, researchers have been exploring various strategies to alleviate computational burdens. Among these, machine learning, especially Physics-Informed Neural Networks (PINNs), has emerged as a promising approach for tackling complex physical problems [8-10]. Raissi's groundbreaking work on PINNs adopted the universal approximation capabilities of neural networks to solve partial differential equations (PDEs) and has opened up a new and efficient avenue for addressing lubrication problems[11]. Subsequent studies by Almqvist[12] and Yang Zhao[13] applied PINNs to solve one-dimensional and two-dimensional Reynolds equations, demonstrating their accuracy in predicting oil film pressure distributions and conducting detailed sensitivity analyses of PINN training parameters. Cheng's HL-nets framework[14], which integrates deep learning with cavitation models, facilitated efficient computations of hydrodynamic lubrication. Further advancements by Xi Yinhu[15] and Rom[16] have refined PINN training strategies by introducing adaptive and loss balance mechanisms. However, none of the current PINN-based lubrication analyses consider roughness effects.

Despite the potential benefits of Physics-Informed Neural Networks (PINNs), a significant obstacle known as the "spectral bias" hinders their performance. This bias causes PINNs to favor learning low-frequency solutions while neglecting higher frequency phenomena[17-20]. To overcome this challenge, researchers have explored various input encoding strategies, such as mapping inputs to higher-dimensional feature spaces, which enhances the network's ability to capture multiscale features[21, 22]. Wang's work on multiscale Fourier feature networks demonstrates potential in solving specific Partial Differential Equation (PDE) problems but requires careful selection of the appropriate number of Fourier feature mappings and their scales [23, 24]. However, these methods have not yet been applied in lubrication analysis. Moreover, employing fixed Fourier eigenfrequencies limits their flexibility in addressing problems with variable or uncertain frequency characteristics, which are particularly prevalent in rough surface lubrication.

In response to this issue, the current research proposes a novel approach where feature frequencies are treated as trainable parameters within the Fourier feature embeddings. This method allows the network to dynamically adjust and select the appropriate frequencies for the task at hand, thereby enhancing its adaptability and flexibility. By applying this novel technique to lubrication problems under varying surface roughness and comparing it with the methods proposed by Wang et al.[23], it is observed that significant improvements in both efficiency and accuracy can be achieved. This progress not only offers a new solution to the spectral bias issue in PINNs but also has implications for optimizing surface textures in multiscale lubrication applications.

# 2 Multiscale lubrication neural networks(MLNN)

## 2.1 Hydrodynamic lubrication

To illustrate the MLNN method, this study uses a slider bearing as an example to demonstrate the process of hydrodynamic lubrication analysis for rough surfaces, as shown in Figure 1.

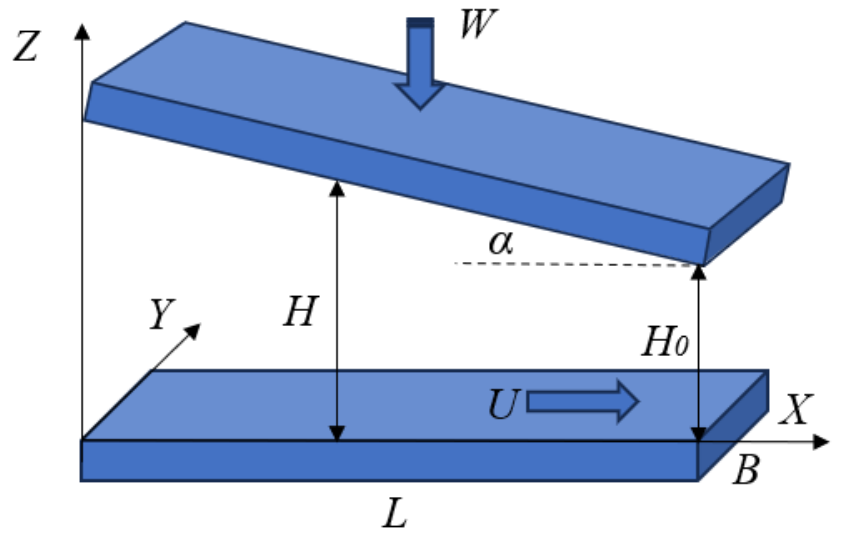

**Fig. 1** slider bearing

In a two-dimensional steady-state, the Reynolds equation for a slider bearing is expressed as:

$$\frac{\partial}{\partial x}\left(h^3\frac{\partial p}{\partial x}\right)+\frac{\partial}{\partial y}\left(h^3\frac{\partial p}{\partial y}\right)=6\eta u\frac{\partial h}{\partial x} \tag{1}$$

Here, "$h$" and "$p$" signify the local film thickness and pressure, respectively. The Cartesian coordinates "$x$" and "$y$" are aligned parallel and normal to the direction of sliding, respectively. The variable "$u$" denotes the relative sliding velocity between the contact surfaces, and "$\eta$" represents the viscosity of the lubricant.

In this study, we conceptualize the lubricant film thickness as comprising three distinct components. The first component, denoted as $h_0(x,y)$, represents the minimum film thickness. The second component, $h_p(x,y)$, reflects the geometric profile of the bearing. The third component, $h_r(x,y)$, accounts for the surface roughness. Thus, the total film thickness at any point (x, y) in the lubrication domain is mathematically expressed as:

$$h(x,y)=h_0(x,y)+h_p(x,y)+h_r(x,y) \tag{2}$$

Upon determining the pressure distribution using a specific numerical method, integrating this pressure over the entire lubrication domain allows for the calculation of the load-bearing capacity, denoted by $W_h$.

$$W_h=\iint_{\Omega} p\,dA \tag{3}$$

The equations were transformed using dimensionless parameters:

$$X=\frac{x}{L},Y=\frac{y}{B},H=\frac{h}{h_0},P=\frac{ph_0^2}{\eta uL} \tag{4}$$

The slider, with a length $L$ and a width $B$, has its upper and lower surfaces moving along the X-axis at a relative velocity $U$. The dimensionless film thickness is given by

$$H(X,Y)=1+\frac{\alpha L}{h_0}(1-X)+H_r(X,Y) \tag{5}$$

where $\alpha$ represents the inclination angle of the slider, and $h_0$ is the thickness of the oil film at the outlet. The hydrodynamic lubrication is modeled by the Reynolds equation. The oil film pressure distribution is derived by solving this equation. The dimensionless form of the Reynolds equation is

$$\frac{\partial}{\partial X}\left(H^3\frac{\partial P}{\partial X}\right)+\frac{L^2}{B^2}\frac{\partial}{\partial Y}\left(H^3\frac{\partial P}{\partial Y}\right)=6\frac{\partial H}{\partial X} \tag{6}$$

A Dirichlet boundary condition , where $p$=0, is applied to the boundaries of the slider.

## 2.2 Multiscale lubrication neural networks architectures

The structure of a Multiscale lubrication Neural Network (MLNN) for solving the Reynolds equation with rough surface, as described by equation (6), is depicted in Fig. 2. The neural network's role is to approximate the solutions $P$ of equation (6). Thus, $P$ is the outputs of the neural network, while the coordinates $X$, $Y$ and $H$ serve as its inputs.

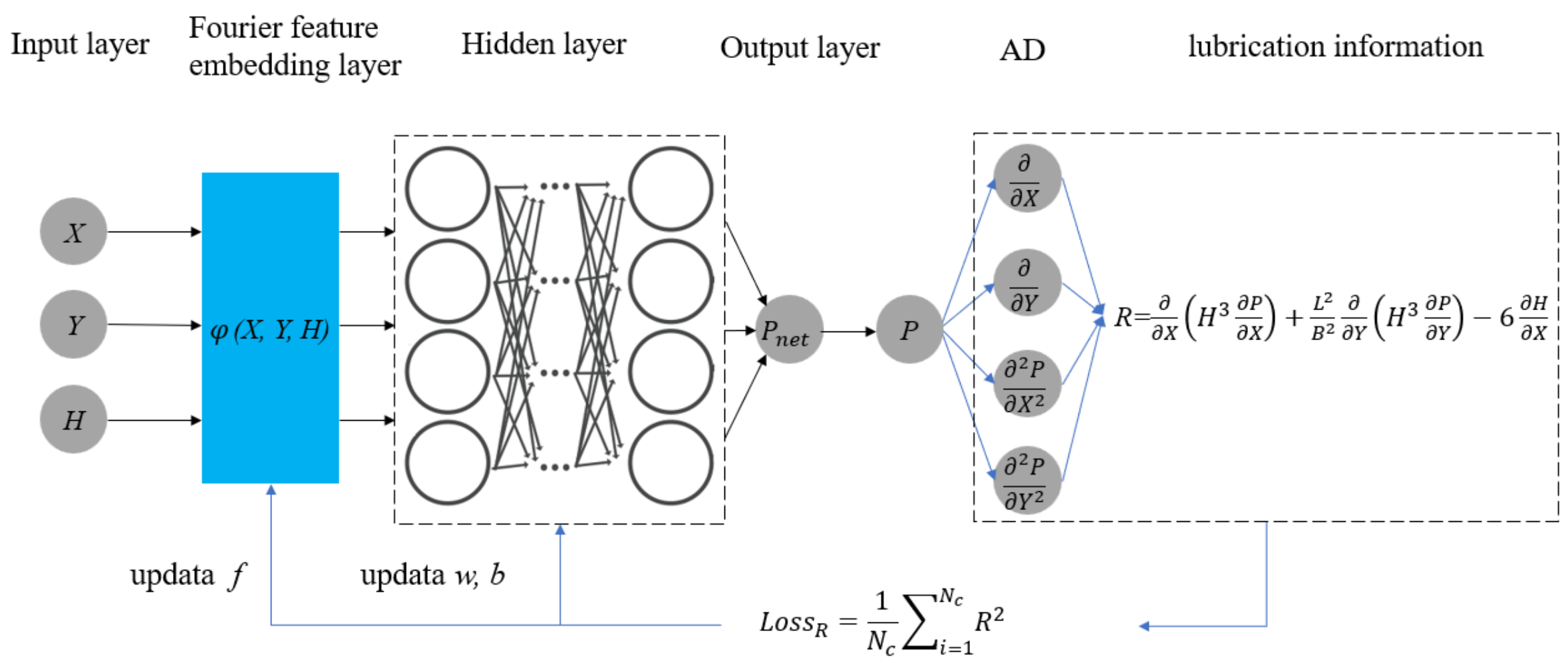

**Fig. 2:** Network structure

Unlike standard physics-informed neural networks (PINNs), the proposed network architecture constructs a Fourier features network using a random features mapping as a coordinate embedding of the inputs, followed by a conventional PINN. The proposed Fourier features embedding can be viewed as a layer with trainable parameters that can be incorporated into any PINN architecture. Compared to the Fourier features network proposed by Wang et al.[23], which uses user-specified frequencies (problem dependent and held fixed during model training), this approach allows for encoding inputs across various frequencies, enhancing the neural network's ability to learn and represent a broader spectrum of functions effectively. Once embedded, the inputs are further processed through a structure consisting of fully connected neural layers.

Mathematically, the transformation of input coordinates into Fourier feature mappings is defined as follows:

$$\varphi^i(X,Y,H) = \left[sin\left(2\pi f^i X\right); cos\left(2\pi f^i X\right); sin\left(2\pi f^i Y\right); cos\left(2\pi f^i Y\right); H\right] \tag{7}$$

For *i=1, 2, ... M,* where $f^{\,i}$ represents the frequency parameters, each sampled from a Gaussian distribution with mean 0 and standard deviation $\sigma_i$ . This encoding strategy significantly enhances the network's capacity to discern and model spatial hierarchies and patterns.

The subsequent processing layers are described as:

$$NN_1^i = \sigma\left(w_1 \varphi^i(X,Y,H) + b_1\right), \qquad for\ i = 1,2 \ldots M \tag{8}$$

$$NN_l^i = \sigma\left(w_l NN_{l-1}^i + b_l\right), \qquad for\ i = 1,2 \ldots M, \qquad l = 2 \ldots L \tag{9}$$

$$P_{\ net}(X,Y,H,w,b,f) = NN_L^i \tag{10}$$

Here $\varphi_i$ and $\sigma$ represent the Fourier feature mappings and activation functions, respectively. The architecture employs weights *w* and biases *b* similar to those in a conventional fully connected neural network, with the addition of a trainable Fourier feature input encoding layer.

When training the neural network, the objective is to minimize the training error regarding prescribed boundary conditions as well as the PDE residual, $Loss = Loss_{bc} + Loss_{R.}$

$$Loss_{bc} = \frac{1}{N_{BC}} \sum_{i=1}^{N_{BC}} (P - P_{bc})^2 \tag{11}$$

Where $P_{bc}$ is the pressure boundary, which is zero in this study, $N_{BC}$ is the training data obtained through the boundary conditions, and Eq.11 ensures that the pressure boundary conditions are satisfied, representing the traditional "soft constraint" form of imposing boundary conditions. According to the "hard constraint" method described in following section in this paper, this term can be removed.

$$Loss_R = \frac{1}{N_c}\sum_{i=1}^{N_c} R^2 \tag{12}$$

$$R = \frac{\partial}{\partial X}\left(H^3 \frac{\partial \hat{P}}{\partial X}\right) + \frac{L^2}{B^2}\frac{\partial}{\partial Y}\left(H^3 \frac{\partial \hat{P}}{\partial Y}\right) - 6\frac{\partial H}{\partial X} \tag{13}$$

In the formula, $N_C$ represents the training point of the equation. By using automatic differentiation technology, the equation residual value $R$ can be efficiently obtained.

### 2.3 Implementation of Hard Boundary Conditions in MLNN

The implementation of boundary conditions within Physics-Informed Neural Networks (PINNs) poses significant challenges, particularly regarding computational efficiency and accuracy. Traditionally, PINNs incorporate boundary conditions through a 'soft' methodology, adding them as terms in the loss function. However, this approach does not guarantee complete adherence to the boundary conditions, affecting both computational performance and solution accuracy. To overcome these limitations, Sukumar [25] introduced a novel method based on R-function theory for the precise application of boundary conditions to neural networks. This method employs the Approximate Distance Function (ADF) to clearly define the boundary, enhancing the enforcement of boundary conditions.

Within the computational domain denoted as $D$ with boundary $\partial D$, the ADF function, $\Phi(x)$, is defined such that $\Phi(x)=0$ for any point $x$ on $\partial D$. For Dirichlet boundary conditions, where the boundary condition $P=P_{bc}$ is prescribed on $\partial D$, the solution ansatz is formulated as:

$$P = P_{bc} + \Phi P_{net} \tag{14}$$

Here, $P$ represents the approximate solution, and $P_{net}$ denotes the output of the neural network.

When the boundary $\partial D$ includes n line segments, the ADF for each segment, $\Phi_i$, ensures that the combined ADFs form $\Phi=\Phi_1 \cup \Phi_2 \cup ... \cup \Phi_n$. Consequently, the ADFs $\Phi_i (x)$ to all partitions of $\partial D$ are calculated, following which the aggregate ADF to ∂D is derived using the R-equivalence operation.

Specifically, when $\partial D$ is composed of $n$ pieces, $\partial D_i$, then the ADF $\Phi$ that is normalized up to order m is represented as:

$$\Phi(\Phi_1, \dots, \Phi_n) = \frac{1}{\sqrt[m]{\sum_{i=1}^{n} \frac{1}{\Phi_i^{\,m}}}} \tag{15}$$

For a detailed discussion on this methodology and its applications, the reader is referred to [25]. The application of this method to lubrication and a comparison with soft boundary conditions are discussed in the author's previous work [26].

**2.4 Training of the MLNN**

For networks with Fourier features, the frequency parameters were sampled from a Gaussian distribution $N(0, \sigma_i)$, where $\sigma_i$ determines the frequency preference of the network's learning process. Consequently, if a network uses only one Fourier feature embedding, the convergence of the frequency components will be slower for all but the preferred frequency determined by the chosen $\sigma_i$. Therefore, it is advisable to embed the inputs using multiple Fourier feature mappings with different $\sigma_i$ values to ensure that all frequency components are learned at the same rate of convergence. In this study, $\sigma_i$ values were selected as 1, 20, and 50, and 30 frequency parameters were initialized from $N(0, \sigma_i)$for each value.

These mappings were then combined into a five-layer fully connected neural network, with each layer containing 100 neurons. Sigmoid activation functions were utilized within the network layers. The weights and biases of the neural network were initialized using the Glorot normal scheme. The parameters were then optimized using the ADAM optimization algorithm with an initial learning rate of 0.01 and a decay rate of 0.005 in MATLAB. Training was conducted for 1000 epochs, with a batch size of 1000. During each epoch, 1000 datasets were propagated through the neural network until all datasets had been processed, updating the network parameters accordingly. Thus, one epoch of training is equivalent to one iteration of the optimization algorithm. 60 collocation points were set up in both $x$ and $y$ directions within the computational domain, such that $N_c = 3600$.

**Table 1** setup of the MLNN and algorithm parameters

| Items | value |
|---|---|
| Fourier feature embedding size | 30 |
| Hidden layers | 5 |
| Hidden layers neurons | 100 |
| Traning epochs | 1000 |
| Mini batch size | 1000 |
| Initial learn rate | 0.01 |
| Decay rate | 0.005 |
| Training point | 3600 |

After completing 1000 training epochs, a comparative analysis was carried out between the predicted outcomes and the solutions obtained through the Finite Element Method (FEM) in terms of pressure distribution, error magnitude, load carrying capacity and peak oil film pressure. The finite element model is solved using the partial differential equation toolbox in MATLAB, where the finite element model contains $60\times60$ nodes, with the same boundary condition of MLNN. These computational experiments were executed on a Lenovo E590 ThinkPad laptop featuring an Intel Core i5 2.3 GHz processor and 8 GB of RAM.

# 3 Results and discussion

## 3.1 Analysis of Results across Varied Surface Roughness Conditions

Case 1: Analysis on a Smooth Surface

The investigation initiated with the deployment of our network on a smooth surface. The pressure distribution plot displays a near-identical curve for both methods, indicating that MLNN can reliably predict pressure profiles similar to the well-established FEM. The absolute error plot reveals predominantly minimal deviation between the FEM and MLNN solutions.

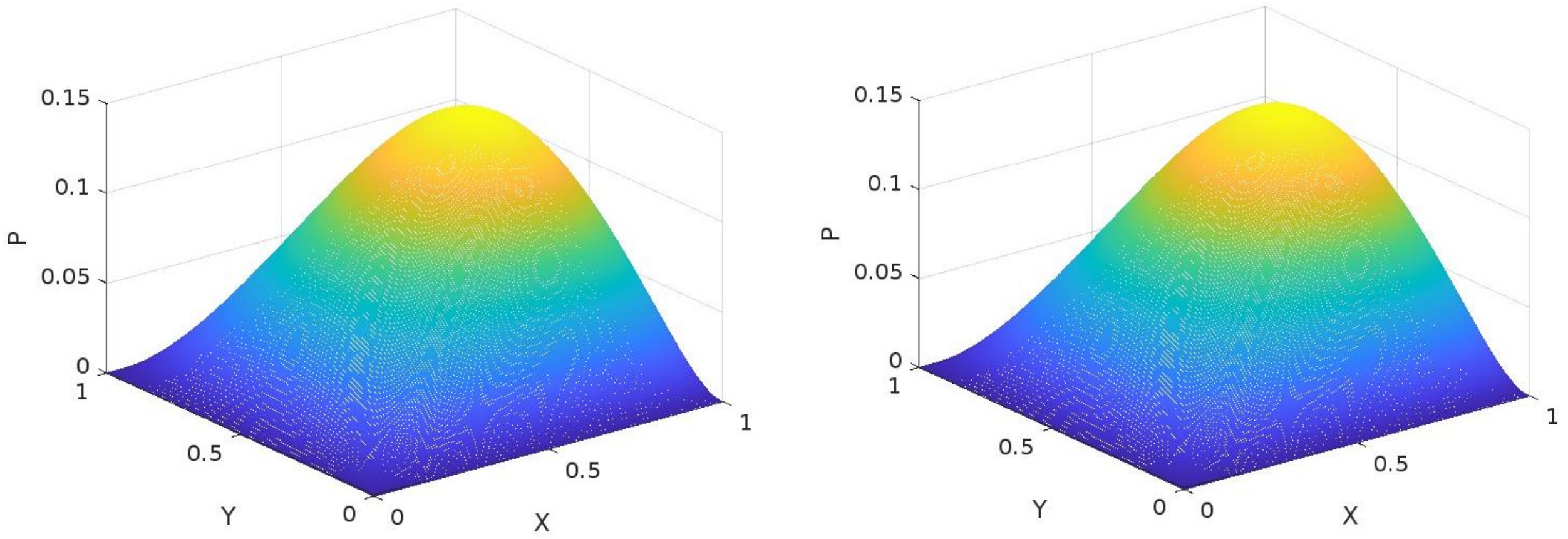

a) FEM b) MLNN

**Fig. 3** pressure distributions with FEM and MLNN methods

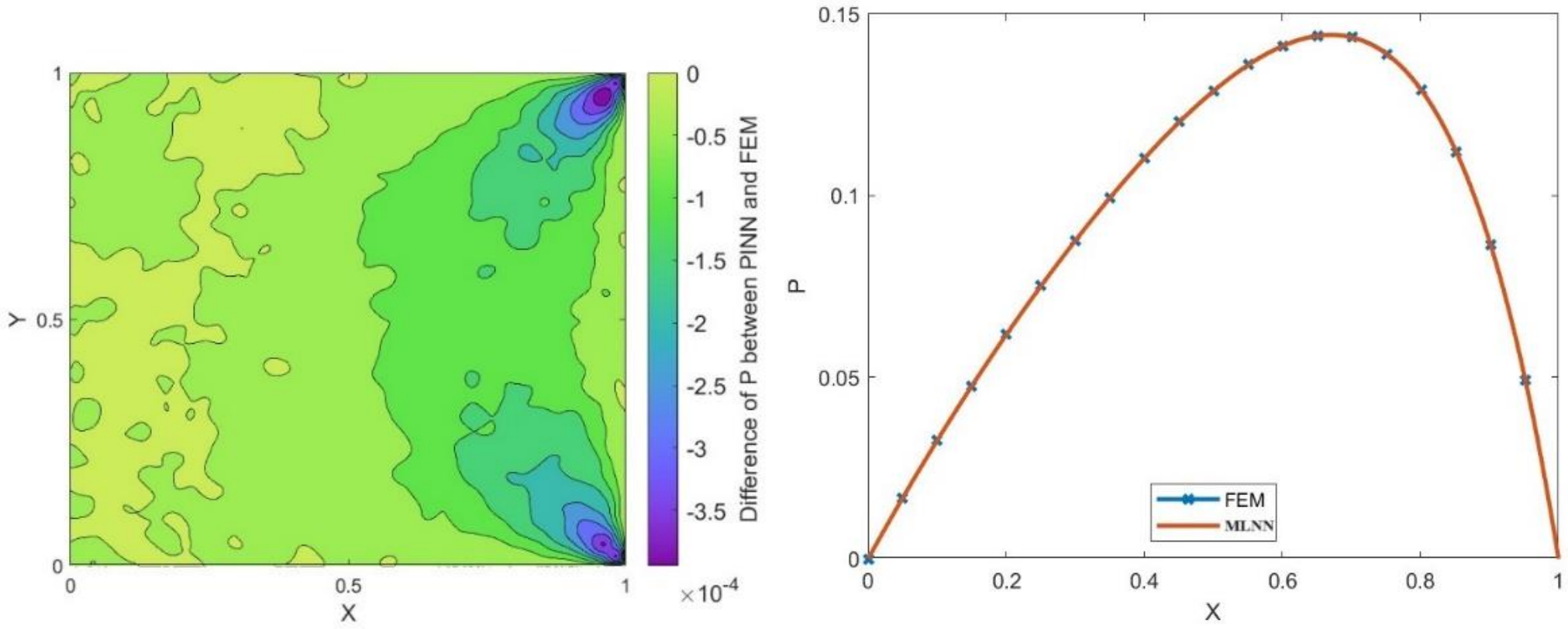

*a) absolute error* *b) pressure distribution at Y=0.5*

**Fig. 4** Comparison of the value with FEM and MLNN methods

Table 1 delineates the maximal pressure and load-carrying capacity computed by both methods. The errors are remarkably negligible, with a maximum pressure discrepancy of only 0.05% and an 0.05% difference in the load-carrying capacity.

**Table 1** Comparison between FEM and MLNN

| | FEM | MLNN | errors |
|---|---|---|---|
| Max. pressure/- | 0.1443 | 0.1442 | 0.05% |
| Load carrying capacity/- | 0.06494 | 0.06490 | 0.06% |

Case 2: Analysis on a Sinusoidal Surface

Subsequently, surface roughness was modeled using a sinusoidal function in Eq. (13) to explore the adaptability of the framework to periodic surface topography.

$$h_r(x, y) = 0.1\, sin(100x) \tag{16}$$

Figure 5 provides a comparison of the pressure distributions obtained by the FEM and MLNN. The pressure distribution shows good agreement over the domain.

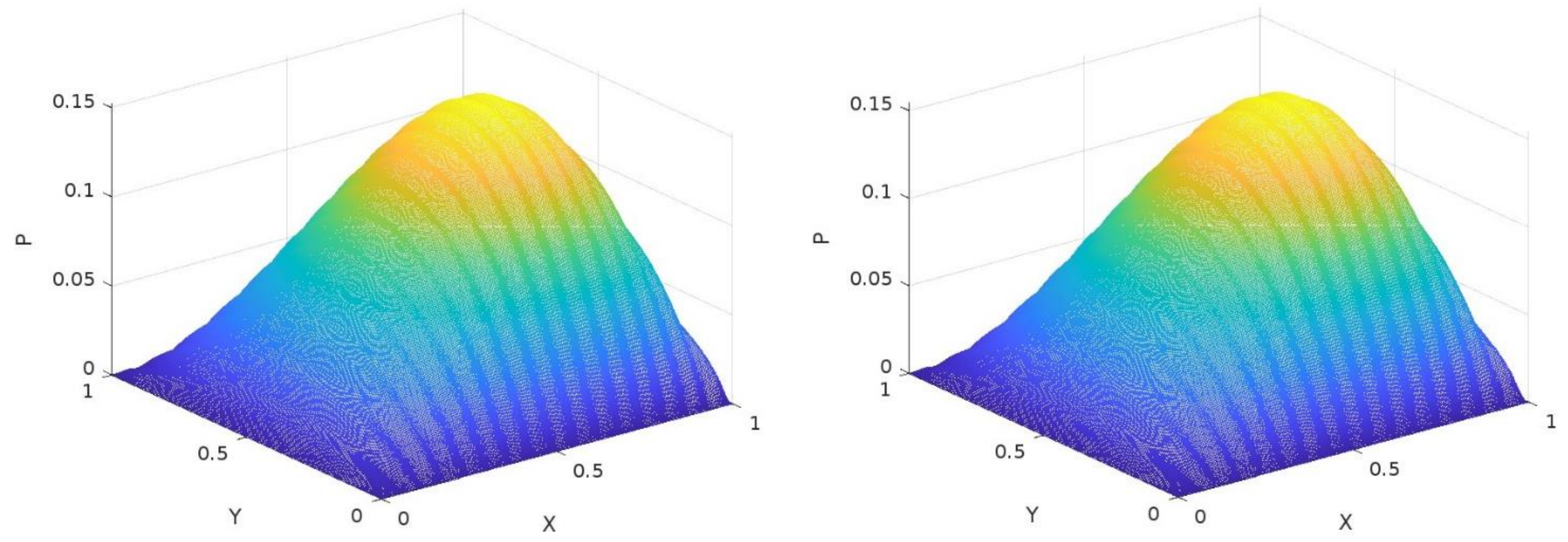

a) FEM b) MLNN

**Fig. 5** pressure distributions with FEM and MLNN methods

In Figure 6a), the absolute error distribution between FEM and MLNN is depicted, showing moderate discrepancies across the domain. Figure 6b) demonstrates the pressure distribution at a specific cross-section (Y=0.5), where the curves from both methodologies closely align, validating the accuracy of the MLNN.

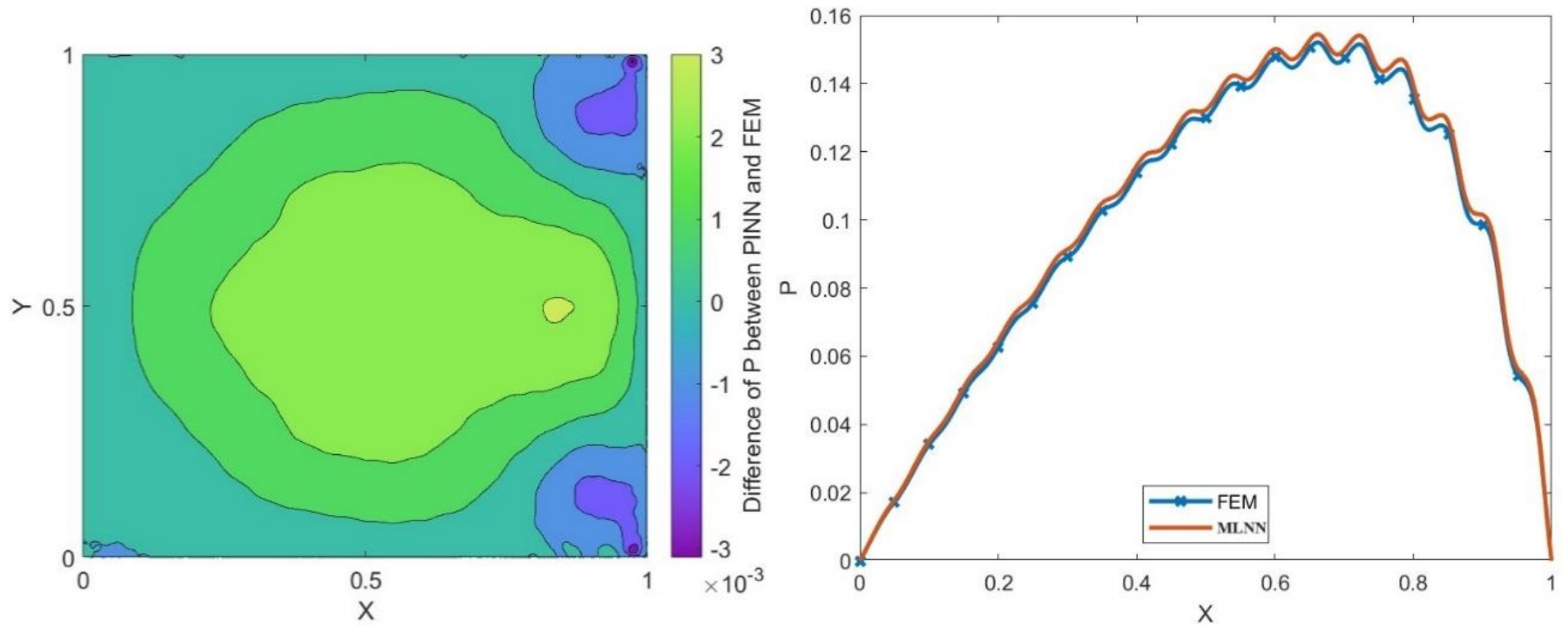

*a) absolute error* *b) pressure distribution at Y=0.5*

**Fig. 6** comparison of the value with FEM and MLNN methods

Table 2 presents a comprehensive evaluation of performance, delineating the maximum pressure and load-carrying capacity calculated by both methodologies. The maximum pressure error was 1.57%, while the load-carrying capacity error stood at 1.77%. Despite these errors being slightly elevated compared to those in the initial case, they still fall within an acceptable range. This reaffirms that even when dealing with the heightened intricacies of a sinusoidal surface, PINN sustains a commendable level of accuracy.

**Table 2** Comparison between FEM and MLNN

| | FEM | MLNN | errors |
|---|---|---|---|
| Max. pressure | 0.1521 | 0.1545 | 1.57% |
| Load carrying capacity | 0.0677 | 0.0689 | 1.77% |

Case3: results for textured surface

In this case, the surface roughness is defined by Eq. (14), wherein A denotes the amplitude of surface roughness, set at 0.2, while $\lambda_x$ and $\lambda_y$ represent the spatial frequency of the roughness, each with a value of 0.02.

$$h_r(x,y) = A\,cos\left(\frac{x}{\lambda_x}\right) sin\left(\frac{y}{\lambda_y}\right) \tag{17}$$

The ensuing outcomes, depicted in Figure 8, delineated the pressure distributions acquired through both MLNN and FEM. These reveal a substantial concordance between the two methodologies. Further insights into the deviations are provided through the absolute error depiction in Figure 9a and the pressure distribution at Y=0.5 in Figure 9b.

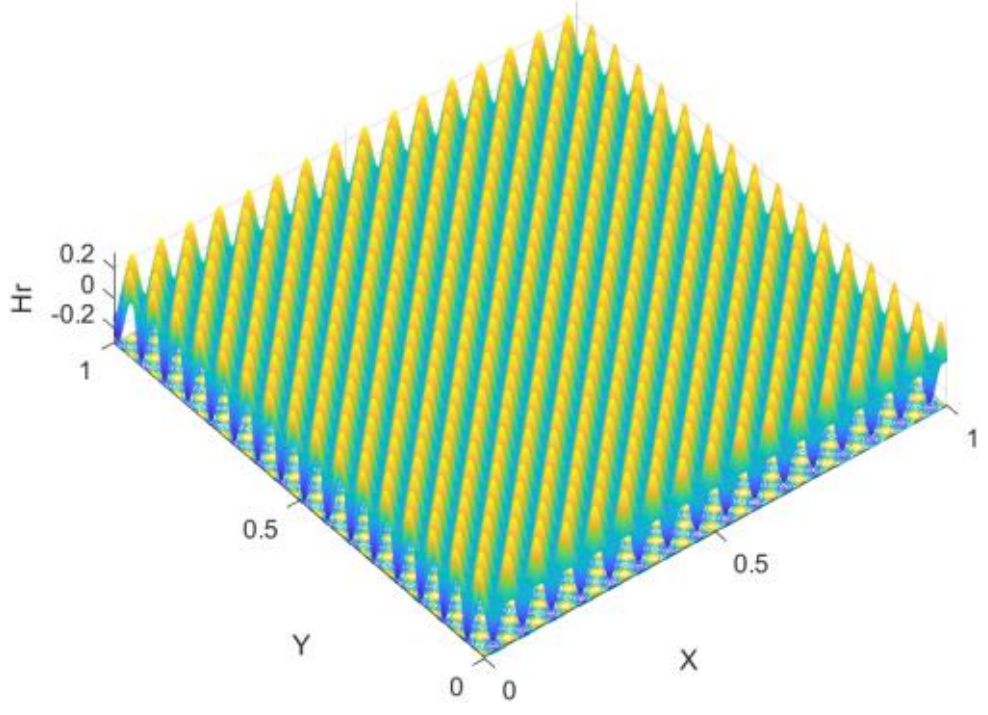

**Fig. 7** rough surface for training

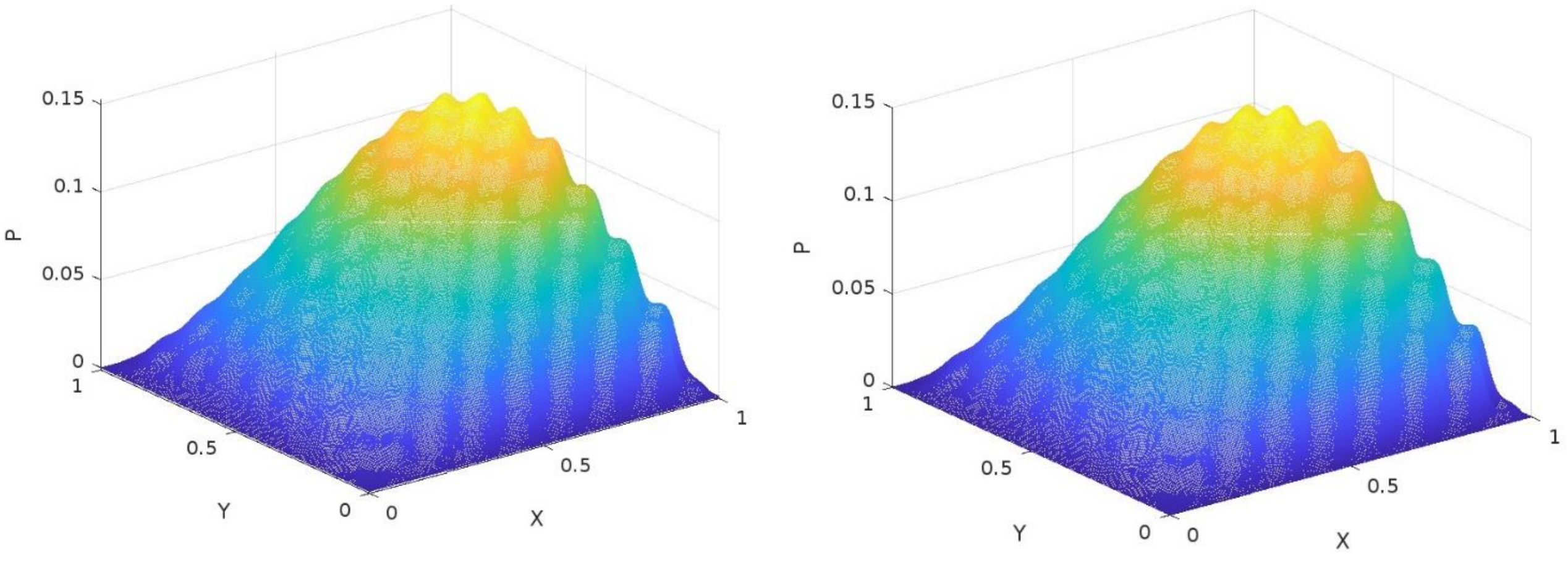

a) FEM b) MLNN

**Fig. 8** pressure distributions with FEM and MLNN methods

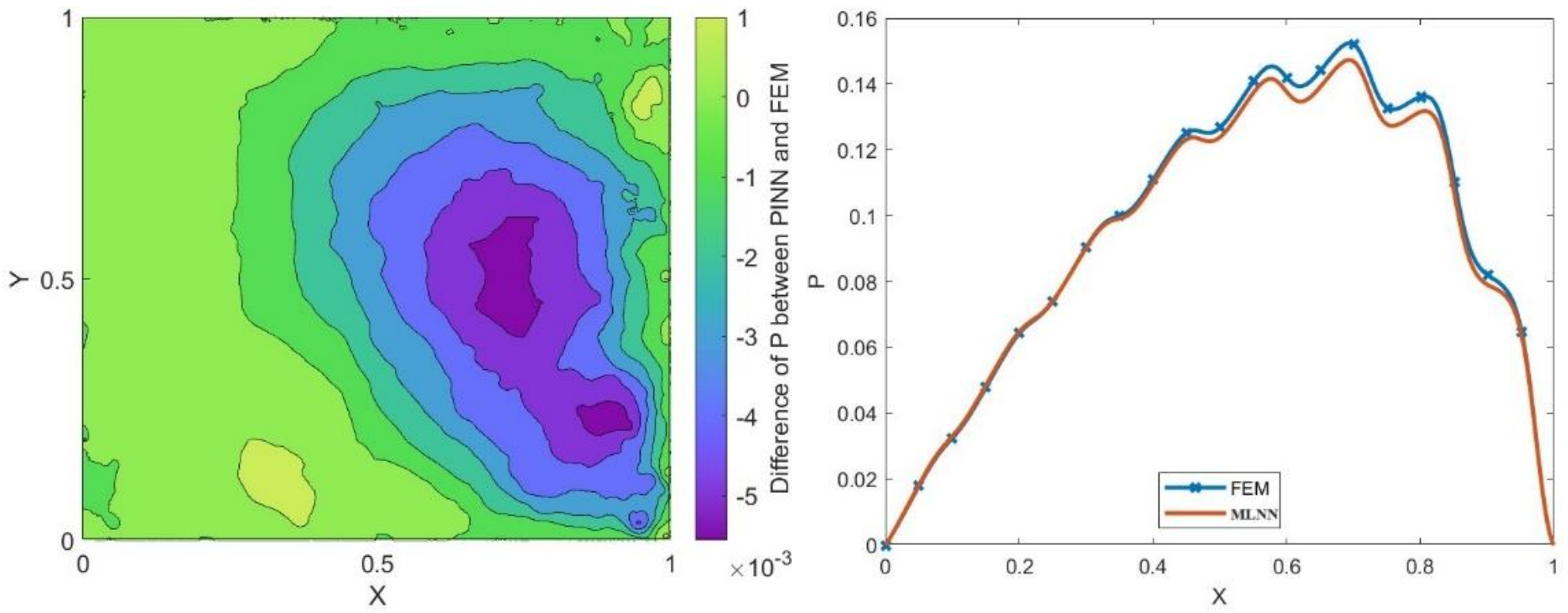

*a) absolute error* *b) pressure distribution at Y=0.5*

**Fig. 9** comparison of the value with FEM and MLNN methods

Table 3 encapsulates the simulation results. The variance in load-carrying capacity is minimal, as MLNN forecasts 0.0644 and FEM predicts 0.0656, exhibiting a scant 1.88% error. The maximum pressure further confirms the capability of the MLNN method with an error of 3.33%, showing a carrying capacity of 0.1474 for MLNN and 0.1525 for FEM.

**Table 3** Comparison between FEM and MLNN

| | FEM | MLNN | errors |
|---|---|---|---|
| Max. pressure | 0.1525 | 0.1474 | 3.33% |
| Load carrying capacity | 0.0656 | 0.0644 | 1.88% |

Case4: Analysis on a Randomly Textured Surface

For this analysis, a Gaussian surface is used as an example to represent the stochastic nature of surface roughness. A comparative evaluation was conducted against the results obtained using the Finite Element Method (FEM) to benchmark the predictive capabilities of our model.

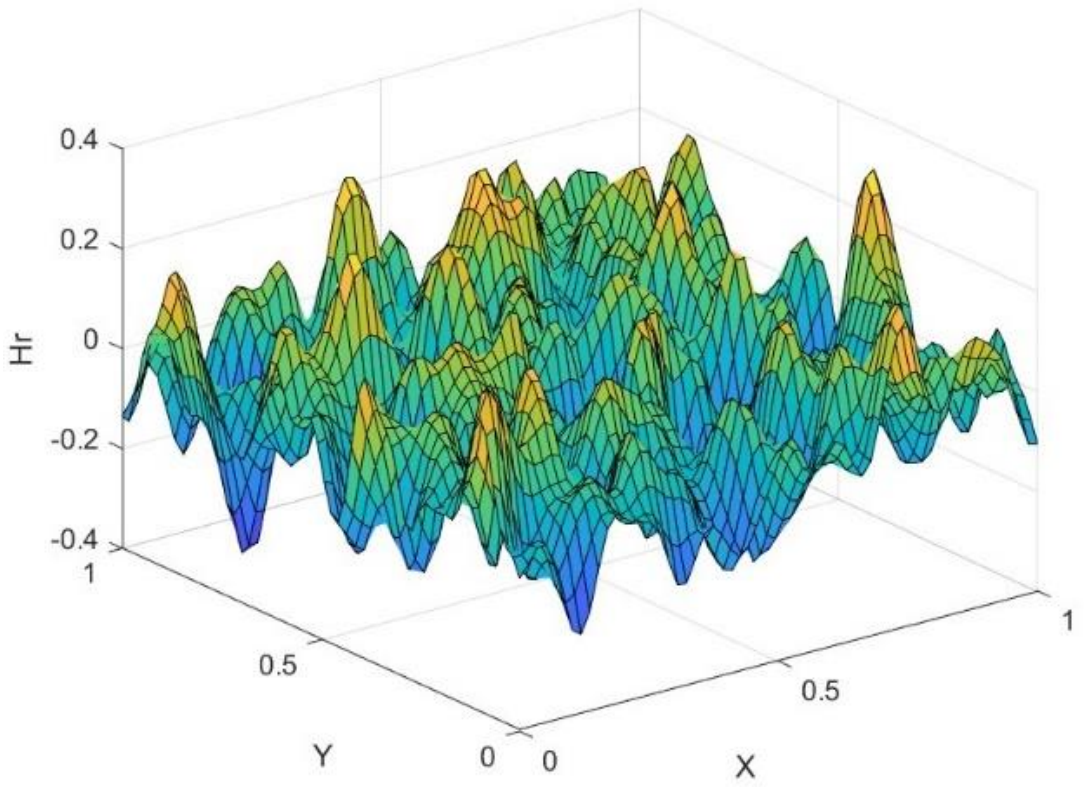

**Fig. 10** rough surface for training

The pressure distributions obtained from both FEM and MLNN are displayed in Figure 11. The comparison reveals a close agreement between the methods, suggesting the neural network's adeptness in capturing the intricacies of the gaussian roughness.

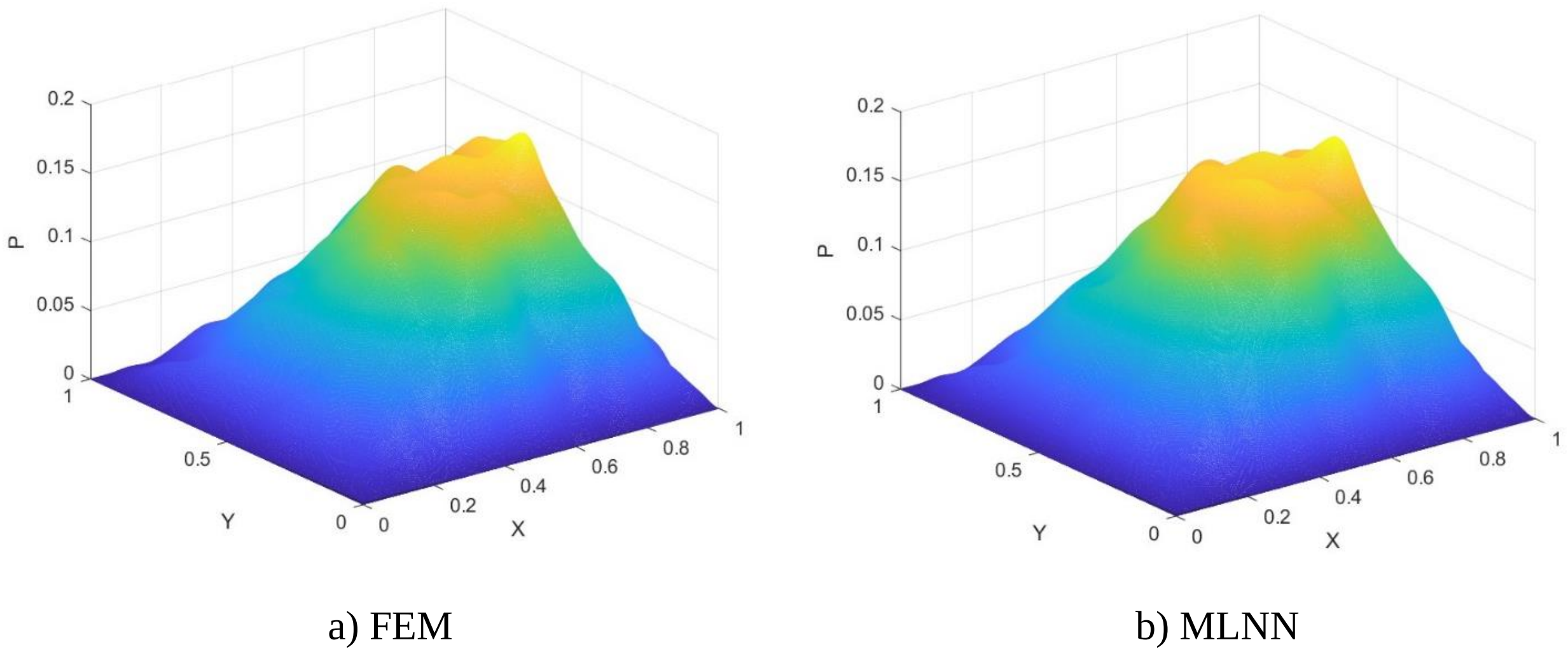

**Fig. 11** pressure distributions with FEM and MLNN methods

Figure 12 offers a more granular view with a contour plot of the absolute error and a graph comparing the pressure distribution at a specific cross-section (Y=0.5). The absolute error distribution graph shows that the deviation is significantly larger compared to the previous examples. The reason for this is mainly caused by the maximum pressure position deviation.

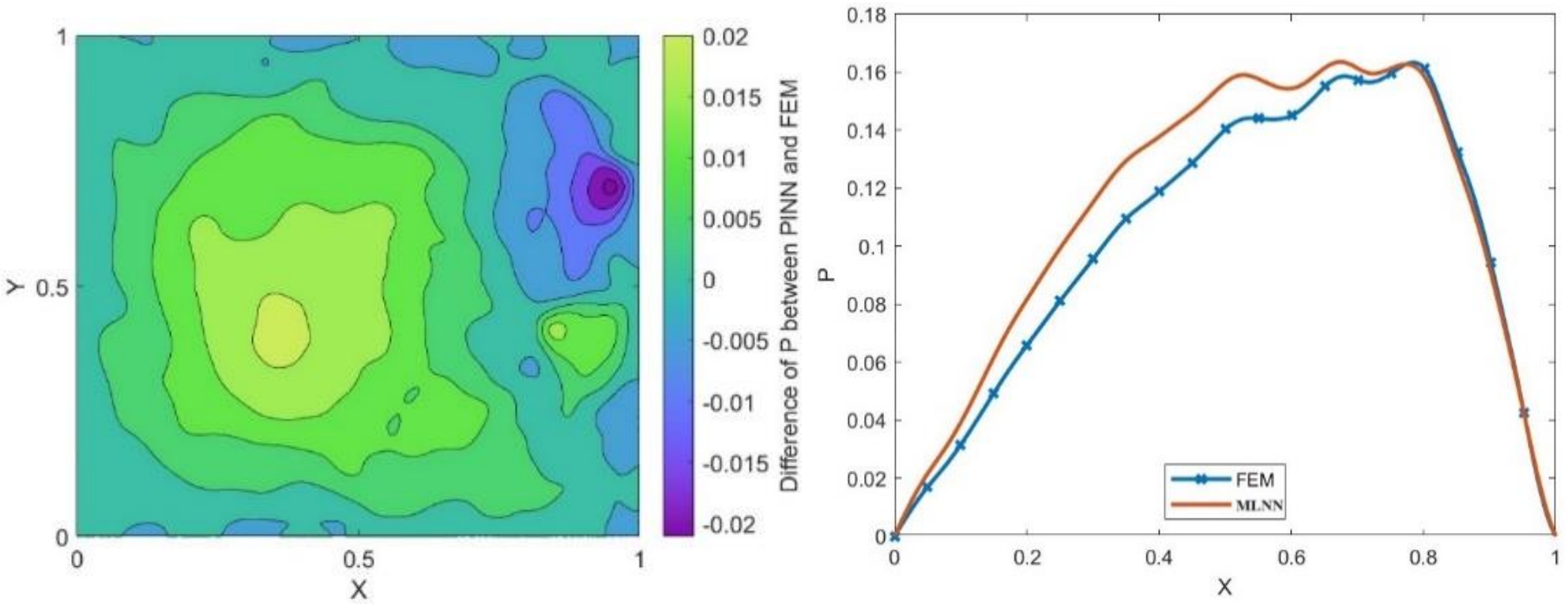

*a) absolute error* *b) pressure distribution at Y=0.5*

**Fig. 12** comparison of the value with FEM and MLNN methods

Table 4 provides a concise summary of the simulation outcomes, presenting the maximum pressure and load-carrying capacity assessed by both the FEM and MLNN. The disparity in maximum pressure values is nearly negligible, with MLNN results showing only a 0.06% deviation compared to FEM. Similarly, the load-carrying capacity exhibits consistency between the two methods, with an error of 8.45%. Although slightly larger than the error in maximum pressure, this deviation is acceptable given that the computation time is less than a second when evaluating the trained neural network.

**Table 4** Comparison between FEM and MLNN

| | FEM | MLNN | errors |
|---|---|---|---|
| Max. pressure | 0.1763 | 0.1765 | 0.06% |
| Load carrying capacity | 0.0689 | 0.0747 | 8.45% |

### 3.2 Effects with frequence learning

To compare the impact of frequency on training, initial frequencies were chosen as follows: $\sigma_1$=1, $\sigma_2$=20 and $\sigma_3$=50. These frequencies remained constant during the training of the network without a trainable frequency.

Figure 13 illustrates the learning dynamics, contrasting trainable and untrainable Fourier Feature embeddings. The trainable Fourier Feature embedding displays a steep learning curve, swiftly descending towards optimization, indicative of an efficient learning process despite initial fluctuations. In contrast, the curve for the untrainable Fourier Feature embedding is less steep, suggesting slower convergence. Remarkably, the trainable network completes the learning process in approximately 13.4 minutes, marginally longer than the 10.33 minutes for the untrainable network. However, this slight increase in training duration is outweighed by the superior accuracy achieved by the trainable network.

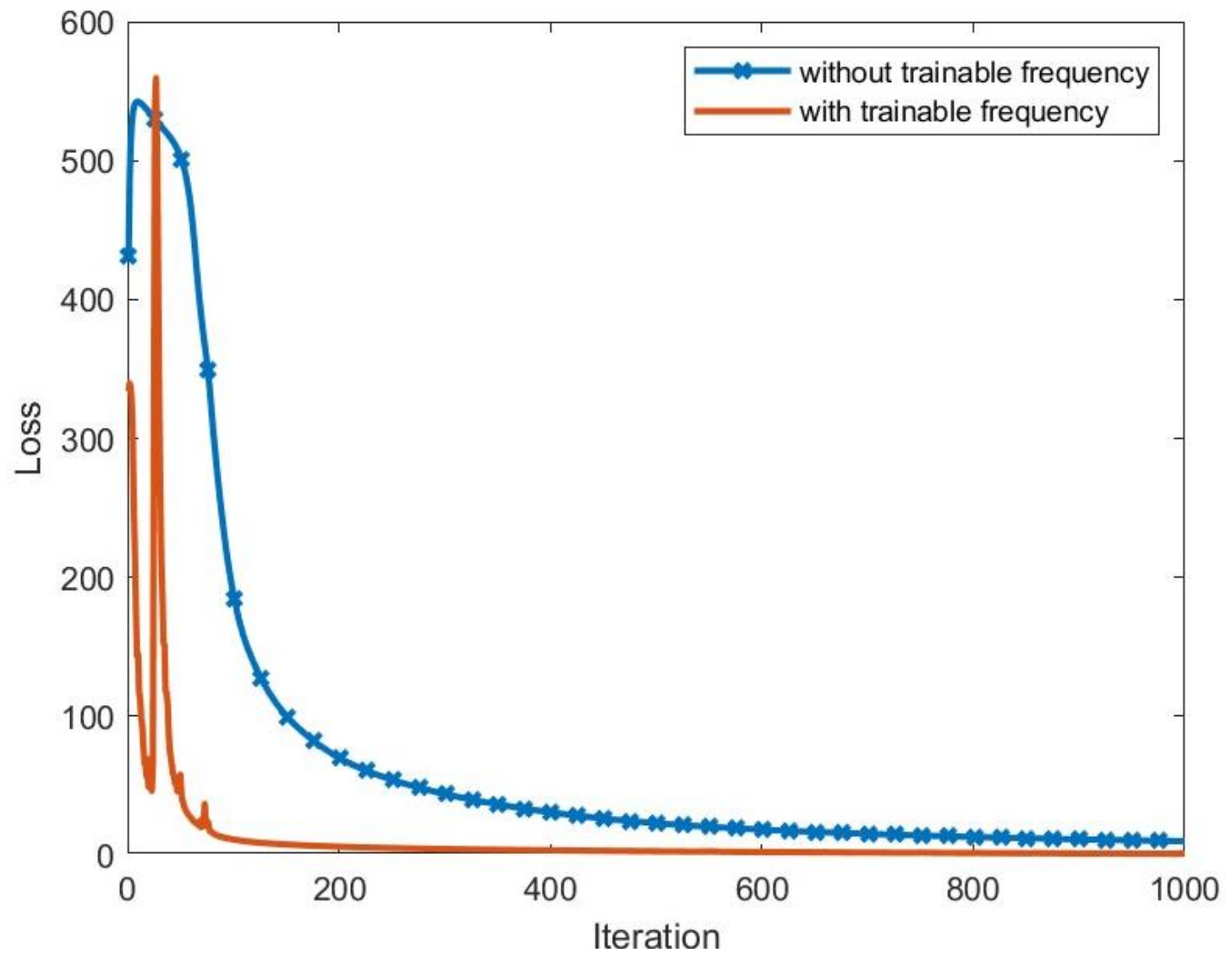

**Fig. 13** Comparison between learnable and traditional Fourier Feature Network

Table 5 provides a quantitative assessment of the two methods, comparing maximum pressure and load-carrying capacity. The errors associated with the MLNN method for maximum pressure and load-carrying capacity are a modest 3.35% and 1.88%, respectively. This stands in stark contrast to the untrainable FFN, which shows errors of 81.11% and 84.94%, correspondingly. These discrepancies underscore the MLNN's superior precision in estimating parameters critical to the tribological performance of lubricated rough surfaces.

**Table 5** Comparison between learnable and traditional Fourier Feature Network

| | FEM | MLNN | errors | FFN | errors |
|---|---|---|---|---|---|
| Max. pressure | 0.1525 | 0.1474 | 3.33% | 0.2762056 | 81.11% |
| Load carrying capacity | 0.0656 | 0.0644 | 1.88% | 0.1213221 | 84.94% |

## 4 Conclusions

In this work, MLNN is used for the first time to solve the Reynolds equation for rough surfaces exhibiting complex multi-scale features. By integrating trainable Fourier feature embeddings into the traditional PINN architecture, this study effectively addressed the spectral bias that has hindered the thorough analysis of high-frequency surface characteristics in lubrication studies. The introduction of a multiscale lubrication neural network framework utilizing trainable Fourier features demonstrated enhanced adaptability to varying surface roughness conditions without requiring prior knowledge of the frequency distribution of the solution. This approach exhibited notable precision and efficiency in lubrication simulations, as confirmed through comparative assessments with finite element methods. Moreover, evaluations indicated that compared to the Fourier feature network with fixed frequency embedding, the proposed model not only boosts computational efficiency but also reduces errors, in predicting pressure distribution and load-carrying capacity, which are crucial for optimizing tribological systems requiring accurate surface interaction modeling.

Future research should focus on further improving the accuracy and generalization performance of this model. Additionally, efforts should be made to expand the model to encompass transient conditions, broadening its relevance to a wider range of engineering challenges. Exploring the integration of other machine learning techniques to enhance lubrication performance prediction under varying operational conditions is also recommended.

**Acknowledgements**

This study was supported by the National Natural Science Foundation of China (No. 52130502) at the National Key Laboratory of Marine Engine Science and Technology. The authors express their sincere appreciation for this work.

**Declaration of competing interest**

The authors have no competing interests to declare relevant to the content of this article.